\pdfoutput=1
\documentclass[final]{clv2025}

\jvol{vv}
\jnum{nn}
\jyear{2025}
\dochead{Last Words} 
\pageonefooter{Action editor: Tal Linzen. Submission received: 10 December 2024; revised version received: 3 March 2025; accepted for publication: 4 March 2025.}
\usepackage{xcolor}
\definecolor{darkblue}{rgb}{0, 0, 0.5}
\usepackage{hyperref}
\hypersetup{colorlinks=true,citecolor=darkblue, linkcolor=darkblue, urlcolor=darkblue}
\usepackage{nert}

\begin{document}

\runningtitle{Natural Language Processing RELIES on Linguistics}

\runningauthor{Opitz, Wein, Schneider}

%\pageonefooter{Action editor: \{action editor name\}. Submission received: DD Month YYYY; revised version received: DD Month YYYY; accepted for publication: DD Month YYYY.}

\title{Natural Language Processing RELIES on Linguistics}

%\author{Juri Opitz\thanks{All authors contributed equally to this work. Email for correspondence: \texttt{\href{mailto:opitz.sci@gmail.com}{opitz.sci@gmail.com}}.}}
%\affil{University of Zurich}

%\author{Shira Wein$^\ast$}
%\affil{Amherst College}

%\author{Nathan Schneider$^\ast$}
%\affil{Georgetown University}

%\author{And Yet Another}
%\affil{Publishing / SPi}

\author{Juri Opitz\thanks{All authors contributed equally to this work. Email for correspondence: \texttt{\href{mailto:opitz.sci@gmail.com}{opitz.sci@gmail.com}.}}$^{,1}$, Shira Wein$^{*,2}$, Nathan Schneider$^{*,3}$}

\affilblock{
    \affil{University of Zurich}
    \affil{Amherst College}
    \affil{Georgetown University}
}

\maketitle

\begin{abstract}
Large Language Models (LLMs) have become capable of generating highly fluent text in certain languages, without modules specially designed to capture grammar or semantic coherence. What does this mean for the future of linguistic expertise in NLP? We highlight several aspects in which NLP (still) relies on linguistics, or where linguistic thinking can illuminate new directions. We argue our case around the acronym \textbf{RELIES}, which encapsulates six major facets where linguistics contributes to NLP: \textbf{R}esources, \textbf{E}valuation, \textbf{L}ow-resource settings, \textbf{I}nterpretability, \textbf{E}xplanation, and the \textbf{S}tudy of language. This list is not exhaustive, nor is linguistics the main point of reference for every effort under these themes; but at a macro level, these facets highlight the enduring importance of studying machine systems vis-\`{a}-vis systems of human language.
\end{abstract}

\section{Introduction}

It is 2025. ChatGPT has been an international sensation for over two years, a focal point in the ongoing public hype, enthusiasm, and concern about ``generative AI.'' NLP conferences are awash in papers about prompting \textit{Large Language Models} that have gobbled zillions of words of text and can produce eerily fluent language (in English, anyway). These models have been refined with corrective feedback on conversational outputs, but (so far as we know) no direct inductive biases about grammaticality or compositional meaning.\footnote{For example, we did not find any such linguistic biases in the reported Llama~3.1 training procedure, though parts of the training targeted code \citep{llama3}.} Like BERT and company an NLP research epoch before---or word embeddings an epoch before that---Large Language Models have elicited both fascination and hand-wringing in scientific circles.

Do these developments spell the end of the relevance of linguistics to NLP?\footnote{We are hardly the first to ask some version of this question: see \cref{sec:background}.} After all, linguistic theories positing elaborate models of syntax are presented as necessary for explaining grammaticality, but are not incorporated into NLP models which nevertheless generate fluent text.

In this piece, we will argue that ideas from linguistics, even if not overtly framed as such, actually continue to underlie much of what we do in NLP. We highlight six facets where linguistics continues to play a major role, and for each facet, point to NLP work which has benefited from (or has been founded on) linguistic knowledge.
We hope our survey will resonate with audiences representing a range of expertise---those with technology-oriented perspectives as well as human language perspectives, and their meeting point within the field of NLP.

\emph{What do we mean by ``linguistics,'' ``NLP,'' and ``CL''?}
Before developing our argument, we need to define our terms (while acknowledging that our definitions will not be perfectly crisp, as disciplinary boundaries are inherently fuzzy).

\textbf{Linguistics} is the study of systematicity and variation in human communication, as transmitted via speech, sign, and writing. Linguists may specialize in how knowledge of forms is organized and deployed to symbolize meanings; how linguistic behaviors emerge, drawing on human cognitive and social capacities; and how language varies over time (ontogeny, phylogeny), geography, identity group, etc.
Linguistics-adjacent fields of language study and practice include language education, translation, rhetoric, communications, and philosophy of language. Below, the term ``linguistics'' covers these (non-AI) ``language fields,'' broadly defined.

\textbf{Natural Language Processing} (NLP) is the field concerned with developing technology for sophisticated computational processing of text, and especially, computational understanding or generation of individual sentences, documents, or conversations (as opposed to drawing inferences about entire collections).\footnote{Text mining, information retrieval, and speech processing are studied in separate research communities, with some overlap. These areas, in combination with prototypical NLP applications such as machine translation and summarization, constitute \emph{language technologies}. The shifting technological landscape may be opening up opportunities for greater unification of these fields \citep[e.g.,][]{chrupala-2023-putting}.} Contemporary NLP research prioritizes empirical evaluation of systems on \textit{tasks}, whether connected with general user applications (like QA, translation, summarization) or more granular and focused on aspects of the language system (like parsing and coreference resolution). 
In recent years, the dominant concern of NLP has been the design, interpretation, and application of pretrained transformer language models such as BERT and GPT. Pretrained models of considerable scale are often called Large Language Models (LLMs).

Today, \textbf{Computational Linguistics} (CL) has two definitions: a broad one (roughly, `computation and natural language') that includes technologically-oriented NLP, and a narrower one that focuses on computational formalization and processing for the end goal of studying how language works. We use ``\textbf{cL}'' for this narrower goal whose chief motivation is to answer questions about language, rather than questions about technology, though methods and sub-questions often overlap between cL and NLP, and consequently both can be found at CL conferences.

\emph{Where would NLP be without linguistics?}
Linguistics is no longer front and center in the way we build NLP systems for practical tasks. Gone are the days when machine translation engines consisted of rules painstakingly crafted by linguists. But there are ways expertise in linguistics continues to be essential in NLP---that is to say, there are ways \textsc{NLP relies on linguistics}.

Imagine what the field would be like \emph{without} linguistics, assuming for the sake of argument that current modeling approaches (word embeddings, transformers, LLMs) had somehow been developed with only a cursory understanding of how language works. Many of the systems might look similar, but the field would look very different. The field would have only the shallowest attention to linguistic analysis (i.e. tasks that focus on representing regularities in linguistic systems beyond the information communicated in a particular application setting). Even for objectives not primarily tied to linguistic analysis, our toolbox would be all the poorer:
\begin{itemize}
    \item \textit{\textbf{R}esources:} We would not have carefully curated datasets such as lexicons and corpora, with an appreciation for variation between languages, dialects, genres and styles, etc. We would not have gold standard annotations of language-system phenomena, only of application-oriented phenomena.
    \item \textit{\textbf{E}valuation:} Not only would we not have gold standard evaluations for linguistic tasks, but for applied tasks, we would also lack expertise for designing effective human evaluations, interrogating automatic metrics, and characterizing the linguistic phenomena that challenge systems (such as anaphora or dialect variation).
    \item \textit{\textbf{L}ow-resource settings:} We would struggle more to understand why approaches that work well for English or French might not work well for Swahili or Arapaho (due to either the lack of data or the features of the language itself). We would not have the knowledge that would allow us to test linguistic inductive biases in neuro-symbolic models for greater accuracy.
    \item \textit{\textbf{I}nterpretability and \textbf{E}xplanation:} It would be harder to develop and test hypotheses about how black box systems such as LLMs process language across domains, and we would lack appropriate metalanguage for describing many observed patterns.
    \item \textit{\textbf{S}tudy of language:} Classic cL tasks such as parsing, coreference resolution, and textual entailment would not exist within our community. An NLP devoted purely to commercial technology would also be indifferent to goals of scholarly or community-driven linguistic work, such as documenting endangered languages.
\end{itemize}

\noindent In what follows, we give a sampling of the ways linguistics continues to play a role in NLP, organized under the mnemonic ``RELIES:'' resources (\cref{sec:resources}), evaluation (\cref{sec:evaluation}), low-resource settings (\cref{sec:l}), interpretability\slash explanation (\cref{sec:explain_and_interpret}), and study of language (\cref{sec:study_of_language}). It is worth emphasizing that these categories and the literature review are meant to be illustrative, not exhaustive; that we are speaking at a macro level; and that we are not claiming that linguistics is the only or necessarily the single most important source of expertise for working with language data and systems. We conclude by discussing pertinent challenges for the NLP community which involve linguistic expertise (\cref{sec:conclusion}).

\section{Background}\label{sec:background}

Today it is not a given that theories and representations from linguistics will form a direct foundation of NLP technologies; it is possible to do research in NLP without traditional training in linguistics. As was the case in the age of the ``statistical revolution'' of NLP \citep{johnson-2009-statistical,uszkoreit-2009-linguistics,church2011pendulum}, deep learning was heralded as a `tsunami' \citep{manning-15-tsunami} and a way to model language `from scratch,' rather than via features reflecting linguistic expertise \cite{collobert-nlp-from-scratch}. Milestones in the neural era included the introduction of word embeddings \cite{NIPS2013_9aa42b31_word2vec} and then pretrained language models \cite{howard2018universal}, with the well-known inflection points of ELMo and BERT \cite{peters2018elmo,devlin-etal-2019-bert} in 2018\slash19 and ChatGPT \cite{openai} in 2022\slash23. The recent past has seen LLMs with billions of parameters that can be accessed by a wide range of users through a natural language interface (`prompting'). Do these developments prove the `bitter lesson' \cite{sutton2019bitter} that language technologies are most effectively developed solely by funneling data on a massive scale into rather general machine learning architectures? What role does linguistics play nowadays in NLP? 

We are not the only researchers to reflect on these apparently tectonic shifts. \citet{ignat2023phd} collect interesting topics for PhD students, particularly evaluation, and \citet{saphra2023first} hypothesize a cyclical historic model that then would suggest that familiar problems will resurface. We also see that the relationship between Linguistics and NLP (as well as the relationship between Linguistics and AI) has been a topic of discussion for some time \citep[e.g.,][]{lakoff-78,raskin-85,nirenburg-86,ws-eacl-vvv-09,ws-acl-common-ground-10}. In 2011, an entire collection of papers was devoted to the relationship between linguistics and CL \citep{baldwin-11}. There it was noted that the field of CL encompasses both scientific and engineering perspectives \citep{kay-11,johnson-11}; and that even with machine learning algorithms, long-tail (rare) phenomena celebrated by linguists are likely to challenge systems \citep{kay-11,levin-11,steedman-11}, 
and claims that NLP solutions will work well for any language need to be carefully evaluated in light of 
linguistic diversity \citep{bender-11} and low-resource conditions in small languages \citep{bird-11}.

In what follows, we offer a contemporary take on how linguistics has enduring, practical relevance for NLP research, under the mnemonic ``RELIES.''

\section{Resources}
\label{sec:resources}

The empirical paradigm that dominates NLP today requires language datasets for training and evaluation.
Such resources are supported by various degrees of linguistic knowledge---ranging from proficiency in a language to formal training in linguistics. 

\paragraph{Resources for general NLP tasks} By ``general NLP tasks'' we mean tasks that closely relate to applications in widespread demand, such as machine translation (MT), summarization, and sentiment classification. Empirical study of these task requires corpus resources. Even if we could now do without some of these resources in the \emph{training} of NLP systems, they remain highly relevant for testing and studying systems.

The \textbf{selection and curation of data can be informed by language expertise}, even if the data is raw text. When developing a dataset, we strive for reflection of language diversity, variation among dialects\slash speakers, genre variation, code-switching, and (if not target of a specific RQ, probably) reduced social biases or problematic content (e.g., hate speech). Sensitivity to linguistic factors is particularly important because of the wide variability of human language and the subjectivity inherent in certain labeling problems, like sentiment and toxic language \citep{sasidharan-nair-etal-2024-exploring,abercrombie-etal-2023-temporal}. Specific data selection techniques may involve shallow linguistic heuristics such as frequency counts, or linguistically advanced techniques that increase and validate the diversity and language phenomena coverage of the data \cite{dryer1989large, rijkhoff1993method}. For example, \citet{ravichander-etal-2022-condaqa} select data with sufficient \textit{prepositions} and \textit{complementizers}, and \citet{ponti-etal-2020-xcopa} inform their data selection with a \textit{language typology index} \cite{littell-etal-2017-uriel}. The BabyLM challenge \citep{warstadt-etal-2023-findingsbaby,hu2024findingsbaby} selects training and testing data according to cognitive linguistic plausibility,  addressing fundamental questions in the learning process of LLMs.

In the annotation process, \textbf{advanced linguistic knowledge of annotators} has been of value, and people skilled at language analysis can ensure meaningful evaluation of NLP systems. For example, \citet{freitag-etal-2020-bleu} observe that the quality and diversity of MT references improves when they are paraphrased by linguists. Additionally, only professional translators (rather than crowd-workers) lead to correct MT system rankings \citep{freitag_etal_2021_mqm}. For consistent sentiment annotation, \citet{sentiment} highlights the importance of understanding linguistic discourse phenomena. Since not all tasks may be suitable for crowdsourcing without extensive experimentation, linguistically trained annotators can help reduce annotation cost \citep{gillick-liu-2010-non}.

Finally, \textbf{designing and coordinating} annotation protocols, especially via crowdsourcing, requires sensitivity to linguistic issues in order to craft guidelines and processes that produce a reasonable level of coherence across the annotations, and to recognize where disagreements can be explained by different perspectives \citep{workflow_crowdsourcing,plank-22,wein-etal-2023-follow,prabhakaran-etal-2024-grasp}.

\paragraph{Resources with linguistic annotations} These encompass datasets of various sorts of grammatical and semantic structures. Linguistic knowledge and awareness is, of course, needed for designing analytical frameworks, and also for conducting annotation and verifying model output.

Two examples of linguistically detailed, resource-oriented frameworks prominent in current NLP research are Universal Dependencies, with morphosyntactic annotations for over 150 languages \citep[UD;][]{nivre-etal-2016-universal, nivre-etal-2020-universal}, and Abstract Meaning Representation, which is a framework for describing semantic graphs that has datasets in several languages \citep[AMR;][]{banarescu2013abstract,wein_coli}. UD and AMR are \textit{applied linguistic theories} \citep[in the sense of][p.~275]{raskin-85}.
Research in and with linguistic annotation frameworks continues apace, with numerous recent workshops in the ACL community focused on grammar and semantics \citep{cxgsnlp-2023-international, tlt-2023-international, dmr-2023-international, udw-2023-universal, depling-2023-international}. The AMR framework alone is featured in over 450 papers,\footnote{\url{https://nert-nlp.github.io/AMR-Bibliography/}} and numerous applications \citep{wein-opitz-2024-survey,sadeddine-etal-2024-survey}.

Generally, linguistically-annotated corpora and NLP models contribute to studying questions about language (\cref{sec:study_of_language}). However, for some major NLP applications (like MT and QA) in high-resource settings, the future of explicit linguistic representations remains to be seen, with mixed results from neuro-symbolic models \citep{hamilton2022neuro,shwartz-fruitless-2023}. On the other hand, we consider the utility of such formal representations for evaluating and interpreting models \citep[e.g.,][]{xu-etal-2021-dynamic, opitz-frank-2022-sbert, fodor2024coling} to be an exciting path (\textbf{E}valuation, \cref{sec:evaluation}; \textbf{I}nterpretability, \cref{sec:explain_and_interpret}).

\section{Evaluation}
\label{sec:evaluation}

We illustrate where linguistics is useful to successfully \textit{evaluate} an NLP system, that is, to describe the degree to which it conforms to our expectations. For tasks that take language input or generate language output, it is important to measure correctness and robustness at the language-system level, for which appropriate evaluations (quantitative and qualitative) are informed by linguistic expertise.\footnote{A similar sentiment was expressed by \citet{levin-11} in the statistical NLP era: ``language technologists [ought to] understand the object of study, human language...in order to understand where current methods are falling short, we as a field need to understand the data'' (p.~19).}

\paragraph{Gold standard evaluation} Linguistic resources, particularly so-called \textit{gold standard} annotations, play an important role in the evaluation of NLP technologies. They also contribute to the \textbf{benchmarking of LLMs and ``AI systems''} \citep{tenney2018what, pimentel-etal-2020-information}: Many of the tasks packaged in the \textit{Beyond the Imitation Game Benchmark (BIG-bench)} \citep{srivastava2022beyond} are linguistic tasks, such as bridging resolution and detecting common morphemes. Another kind of benchmark tests linguistic inference from just a few examples in puzzles \citep{sahin-etal-2020-puzzling, chi-etal-2024-modeling}, a format conducive to low-resource languages without established annotation conventions \citep{bean2024lingoly}.

Linguistic resources are also crucial for \textbf{system diagnostics} (including error analysis), which is a part of evaluation, but does not necessarily have the aim to produce a ``global'' ranking of AI systems. Instead, diagnostic methods help find more targeted improvement perspectives, with possible relevance even to broader society, since linguistic phenomena are often contingent with issues of societal interest, like gender biases in coreference \citep{rudinger-etal-2018-gender} or QA \citep{parrish-etal-2022-bbq}.\footnote{\citet{raji2021ai} generally argue for more targeted evaluation methods to track ``AI progress.''} Overall, linguistic diagnostics can come in many forms; here we give a small sample. The `CheckList' \citep{ribeiro-etal-2020-beyond} tests NLP systems in various aspects, using linguistic features (negation, part of speech, etc.)\ to build templates. \citet{song-etal-2022-sling} assess LLMs on Chinese linguistic phenomena, and \citet{parcalabescu-etal-2022-valse} employ linguistic aspects to study (computer) vision and language systems. \Citet{moore-2009-computational} discusses examples of challenges that  require linguistic knowledge to unpack, such as structural parallelism and translation of ``WH'' questions. Automated diagnostics can be achieved, e.g., through targeted measurements on parsed semantic structures \citep[][calling for work on the intrinsic NLP task of parsing]{lo-wu-2011-meant, opitz-frank-2021-towards, fan-etal-2023-evaluating}, or stratifying evaluation data by semantic complexity \citep{antoine-etal-2024-linguistically}. Then there is a large area of probing systems with linguistic structures \cite{tenney2018what-probing1,rama-etal-2020-probing, pimentel-etal-2020-information-probing3, starace-etal-2023-probing4}, also tailored at certain branches of linguistics, e.g., psycholinguistics \cite{gauthier-etal-2020-syntaxgym, ettinger2020bert}.

\paragraph{Human evaluation and meta-evaluation} The combination of varied \textbf{human evaluation} practices with progress towards highly fluent models has led to diminished reliability of human evaluations: Since generated text has become so fluent, it is harder for humans to distinguish their quality.
As a result, not all human evaluation studies are useful measures of model output \citep{clark-etal-2021-thats,freitag_etal_2021_mqm}, necessitating standardization with regard to design and terminology \citep{howcroft-etal-2020-twenty,belz-etal-2020-disentangling,van-der-lee-etal-2019-best}.
Making use of implicit linguistic knowledge held by native speakers (e.g., judging the fluency of system output) or more explicit linguistic training (e.g.~assessing the syntactic diversity of system output), can serve a crucial role in reliably assessing the state of the field \citep{michael-etal-2016-putting,callison-burch-etal-2008-meta}.

To design approaches for \textbf{meta-evaluation} (assessing and comparing metrics), expertise in linguistics can prove highly useful, since the challenging linguistic phenomena for models and metrics need to be understood in order to be identified. In a WMT metric shared task, \citet{freitag-etal-2021-results} find that most metrics struggle to handle complex semantic phenomena and do not appropriately punish the reversal of sentiment or negation; this is established with test sets that assess specified linguistic features (here: sentiment or negation reversal).

\paragraph{New metrics} Linguistic annotations can help build automatic metrics. Some NLG metrics are built from entailment data\footnote{While entailment annotations may come from non-linguists, the notion of textual entailment was inspired by traditions from logic and formal semantics \cite{dagan_nli,bowman-etal-2015-large}.} \citep{10.1162/tacl_a_00576, xie-etal-2021-factual-consistency, scialom-etal-2021-questeval,steen-etal-2023-little}. Measures for biases, like social biases, have drawn upon the efforts of linguists. For instance, \citet{steen-markert-2024-bias} propose bias evaluation measures from socio-linguistic lexicons (gender, race) and OntoNotes \citep{pradhan-etal-2013-towards}.

\section{Low-Resource Settings}
\label{sec:l}

In NLP and ML, we aim to \textbf{efficiently learn to generalize} to new tasks, domains, and languages, including \textbf{global and historic languages}, where feedback and data is sparse or nonexistent \citep{hedderich-etal-2021-survey}. 

\paragraph{Processing of global and historic languages} 

Linguistics is crucial in situations where the amount of language data is limited, e.g., when developing technologies for languages with a limited amount of available recordings or written data. We can improve generalization performance across an array of LLM models by exploiting linguistic features in training \citep{zhang-etal-2024-hire} or dictionaries for balanced data augmentation  \citep{lu-etal-2024-llamax}; also with benefits to speech processing \citep{kim-etal-2024-audio}. Processing particular types of language\slash particular languages can also pose challenges that require input from linguistics. For examples, \textit{polysynthetic languages} such as those spoken in Canada \citep{gupta-boulianne-2020-automatic, gupta-boulianne-2020-speech} pose problems for automatic speech recognition (ASR): if `word' is taken as the basic unit for recognition (which makes sense for most languages, but not these ones) the out-of-vocabulary (OOV) rate on new data is extreme.

At first glance, \citeposs{tanzer2024a} study may appear as a potential `linguistics-free' avenue for addressing some extremely low resource-scenarios, since it suggests that LLMs could, to a certain extent, translate a language from one linguist's grammar book instead of a corpus. However, the study shows crucial reliance on linguistics: a linguist is required to assess LLM performance, and even the largest LLMs considerably underperform relative to a linguist (unfamiliar with the language) who has read the book. Of course, it is also the case that the fundamental resource was created by a linguist.

Even when more data is available, linguistics can be of value: It can inform data selection in multi-lingual pretraining \cite{ogunremi-etal-2023-mini}, or more language-balanced tokenization \cite{creutz-lagus-2002-unsupervised, limisiewicz2024myte, fusco2024recurrent}.\footnote{Some research also suggests that ``simply scaling up the number of languages [...] in the pretraining is unhelpful'' for generalization \citep{adelani2023sib}, and ``bigger is not always better'' \cite{wilcoxbigger}.}

\paragraph{Resources for endangered languages}\label{par:resources-endangered} 
According to some predictions, half of the world's spoken languages might be on the brink of extinction \citep{UNESCO.org_2023}. Thus, there is scope for NLP tooling to document and revitalize endangered\footnote{Terms like `endangered' and `documentation' are widespread in the literature we are surveying, though we also wish to point at a critique of this terminology, suggesting `reclamation' instead \citep{Leonard:20}.} languages \citep[e.g.,][]{van-esch-etal-2019-future,levow-etal-2021-developing,san-etal-2022-automated}, ideally through collaborations involving community members, linguists, and NLP practitioners. Linguistic expertise is relevant for structurally collecting and documenting language data, as well as helping to distill knowledge about the language into dictionaries and rules. Recent work has brought community members and NLP practitioners together for the purpose of language documentation through provisioning access to speech tools such as ASR \citep{san-etal-2022-automated}, building datasets on online platforms \citep{everson-etal-2019-online, zuckermann-etal-2021-lara}, and analysis of phonetic data \citep{kempton-2017-cross}.

\paragraph{Compute restriction}  
Systems that require less compute, particularly at inference time, are of great interest. Critically, off-the-shelf conventional parsers\slash taggers defend their status as the go-to method in some situations, like for a certain forms of data analysis (e.g., counting the most common nouns), or affordable processing of huge corpora (e.g., from sentence segmentation to OpenIE \citep{angeli-etal-2015-leveraging}).

Effective engineering solutions to efficiency are achieved through quantization \cite{NEURIPS2022_c3ba4962_quant} or low-rank adaptation \cite{hu2022lora} of LLMs. However, soft inductive linguistic biases might be complementary, or an alternative in specific situations. Tailored evaluation metrics could help assess any incurred tradeoffs \cite{zhou-etal-2022-assessing}.

\paragraph{Linguistically sensitive supervision} 
Developers of technologies for under-resourced languages do not only encounter data scarcity, but probably also scarcity of input from native speakers who could help oversee a system's design and application. Indeed, releasing technology in a top-down fashion can be harmful to local communities \cite{bird-2022-local,dogruoz-sitaram-2022-language, bird-yibarbuk-2024-centering}. Thus, when applied with care, \textit{(field) linguistics} can help become aware of the diverse cultural contexts of local language communication situations, resulting in more sensitive  NLP technologies, or the abstention from releasing harmful ones (e.g., when there is no community oversight/feedback).

\section{Interpretability and Explanation}
\label{sec:explain_and_interpret}
 
How can we efficiently reason about, and explain observations of observed language phenomena, language systems and processes, as well as any of their models in the form of NLP systems? We need a \textbf{shared terminology and metalanguage}, and develop \textbf{binding methods} that relate model internals to human-understandable concepts from said terminology. We argue that linguistics fundamentally helps with both aspects.

 As the field concerned with systems of human language, linguistics offers a \textbf{metalanguage}---in the form of jargon and formalisms---for precise characterization of linguistic phenomena. The field of NLP draws on this metalanguage, as can be seen in the literature in observations like ``phonemic representations exhibit higher similarities between languages compared to orthographic representations'' \citep{jung-etal-2024-mitigating} and ``difficult to score [in MT evaluation] are the transitive past progressive, multiple connectors, and the ditransitive simple future I for English to German, and pseudogapping, contact clauses, and cleft sentences for English to Russian'' \citep{avramidis-etal-2024-machine}.

For understanding computational models of language such as LLMs with billions and trillions of parameters, directed \textbf{binding methods} are being developed that learn to relate model internals and processes to specific human-understandable descriptions, that often lie  within our vocabulary of linguistic metalanguage. We will consider a few examples of such methods that particularly use a linguistic notion of binding: \citet{rassin2023linguistic} bind linguistic concepts in V\&L diffusion models with syntax trees and attention maps; \citet{di-marco-etal-2023-study} and \citet{jumelet-zuidema-2023-feature} detect linguistic feature representations in LLMs with prompts; \citet{opitz-frank-2022-sbert} bind parts of neural embedding representations to semantic aspects; \citet{geva-etal-2022-transformer}  view model decisions as constructed from concepts in a vocabulary. Explanations and binding can also adopt an outside, behavioral view \cite{beguvs2023large, behzad-etal-2023-elqa, chang-24}. For instance, \citet{munoz2023contrasting} contrast linguistic patterns in human- and LLM-generated text in morphological, syntactic, and sociolinguistic aspects. Since such methods can be also seen as diagnostic methods, the section on \textbf{E}valuation (\cref{sec:evaluation}) is also relevant. 

On a bigger picture, linguistic metalanguage can help us understand, or establish hypotheses, for what happens in complex NLP systems, and how they differ. An example is the hypothesis that a pretrained LM can be viewed through the lens of a classical NLP pipeline, which has received arguments for and against \citep{tenney-etal-2019-bert, de-vries-etal-2020-whats,niu-etal-2022-bert}. The learning trajectory of LLMs can be traced linguistically, e.g., noting the point of syntax acquisition \citep{chen2024sudden}. LLMs have also been interpreted as ``models of varieties of language'' \citep[a sociolinguistic notion,][]{grieve2024sociolinguistic}.

Building a bridge to \textbf{R}esources (\S\ref{sec:resources}), we can use \textit{data sets} that elicit linguistic phenomena to ``open the black box'' of LLMs down to the level of single neurons, as exemplified by \citet{niu2024what}, who study neuron-level knowledge with the BLiMP corpus of linguistic minimal pairs \citep{warstadt-etal-2020-blimp-benchmark}. With similar means, \citet{wang-etal-2024-beyond-agreement} show that BERT-like models focus on sentence-level features, whereas LLMs such as GPT or Llama are sensitive to conventions, language complexity, and organization. Experiments with complex and simple language structures can illuminate learning environments for LLMs \citep{qin2024sometimes}.

\textbf{Linguistically-agnostic interpretability} methods can be mechanistic, i.e., formalizing the algorithms learned by neural networks \cite{pmlr-v139-weiss21a}, or based on attributing a prediction to inputs via integration \cite{integrated_gradient}, classifiers \cite{ribeiro2016should-lime}, or Shapley values \cite{shapley1951notes}. However, still, for interpreting and evaluating the results of such methods, we rely on linguistic analysis \cite{schuff-saliency-methods-need-lingustic-interpretation, feldhus-etal-2023-saliency, moeller-etal-2024-approximate}, to verify that an interpretability method ``outputs meaningful explanations'' \cite{pmlr-v130-mardaoui21a} or ``emphasize[s] specific types of linguistic compositions'' \citep{kobayashi2024analyzing}. This means that linguistic analyses are used to validate interpretability methods whose algorithms do not explicitly incorporate such representations.

Finally, at the thousand-foot view, we see ideas from linguistics and adjacent fields take center stage in debates about how to \textit{interpret} what NLP models are capable of representing---like how to define machine `understanding' \citep{dunietz-etal-2020-test,ray-choudhury-etal-2022-machine}, and specifically, whether grounding is required for a model to capture meaning \citep[e.g.,][]{bender-20,merrill-21,pavlick-23}.

\section{Study of Language}
\label{sec:study_of_language}

Our last facet highlights linguistics (and related areas) as the application domain. That is, those who study language, but are not necessarily computational linguists themselves, are a user base that can motivate NLP tasks and tools.\footnote{Unlike the previous categories, which highlight ways that linguistics contributes to NLP, we view this category as bidirectional: language study motivates research towards developing NLP tools which can then contribute to the study of language. To be clear, we regard the entirety of cL---spanning formal, descriptive, distributional\slash statistical, and experimental inquiries---as part of linguistics. (This includes the development and use of frameworks and resources of the sorts discussed in \cref{sec:resources}.) These inquiries contribute to linguistic understanding by advancing accounts of linguistic diversity and the precision and robustness of theories. Naturally, NLP models and representations are part of the methodology of this research (in particular, the potential relevance of LLMs to broader linguistics and cognitive science has been widely debated: see \citet{futrell-25} for a review).
For purposes of this section, though, we focus on specialized NLP software that can be applied to language studies}, not NLP models and representations generally. These \emph{language-system-focused} applications can be distinguished from most contemporary NLP applications, which seek to make \emph{content} more accessible through language (for example, by translating it, summarizing it, or reasoning about it).

\textbf{Classic cL tasks} like parsing have applications to corpus linguistics: in order to study a linguistic pattern, it is often important to query not just strings, but with grammatical abstractions like tags, dependencies, and phrases. This was not the original goal of parsing researchers: parsing was a cL topic before the days of substantial digital corpora, and was studied to examine computational capacity to model natural language grammar, highlighting issues like formal expressive power \citep{steedman-11}. Now, the tools produced have found a secondary application in corpus linguistics, with parser outputs accessible through corpus search engines \citep{resnik-05,fangorn,Ben:Gho:Bal:Dri:12,hundt-12,annis,grew,kulick-22}.\footnote{Search in parsed corpora may also be motivated by information extraction applications \citep{shlain-20}.} The application to corpus linguistics motivates new angles on parsing research (e.g., syntactically searching a large text collection without having to pre-parse it in its entirety, or in a way that accounts for parser uncertainty). 

\textbf{Documentary and historical linguistics} also motivate and contribute to NLP advancement. In documentary and historical linguistics, data may be sparse, fragmented, primarily in image or audio form (without transcriptions), and lacking a standard orthography; the basic grammar of the language may still be a mystery; and the language may have few or no living speakers. (The case of endangered language documentation was discussed above in \cref{sec:l}.) The exigencies of such settings call for considerably more noise-tolerance and interactivity than run-of-the-mill NLP tasks. Pure-text tasks like normalization \citep{robertson-goldwater-2018-evaluating}, as well as signal processing capabilities like OCR for historical documents \citep{berg-kirkpatrick-etal-2013-unsupervised} or low-resource speech processing \citep{duong-etal-2016-attentional,anastasopoulos-etal-2017-spoken,liu-etal-2022-enhancing}, are important here. Algorithms and software infrastructure to assist humans engaging in multiple tasks collaboratively (elicitation, transcription, annotation, discovering grammatical generalizations, and developing usable resources for scholars and community members) in limited-data multimodal conditions are needed, and there is a long way to go before they can be used seamlessly by non-computational linguists \citep{gessler-2022-closing,moeller-arppe-2024-machine}.

\textbf{Language teaching}, as studied in the field of Applied Linguistics and practiced in classrooms as well as other modes of instruction and technological support, is another area ripe for further engagement with NLP. Much of the research in the NLP community has been formulated in narrow tasks like grammatical error correction (as a generation task) or essay scoring (as a regression task) \citep{burstein-99,massung-16,wang-21,klebanov-22}. Applications with richer contexts---for example, technologies that would monitor student progress over time and deliver adaptive pedagogical experiences---are perhaps more difficult to study, but could enhance student learning more holistically \citep{cui-23,qian-23,glandorf-25,bang-24}.

Finally, we note the potential for NLP to embrace applications facilitating language-focused study in fields beyond linguistics proper, including literature \cite{lit_language}, education, law \cite{law_language}, communications \cite{comm_language}, translation studies \cite{translate_language}, history \cite{piotrowski2012natural}, argumentation \cite{cohen-1987-analyzing}, and lexicography \cite{lex_language}. In the NLP community, these fields are associated and developed with active workshops, e.g., LaTeCH-CLfL (CL for Cultural Heritage, Social Sciences, Humanities and Literature---8th iteration in 2024) \citep{latechclfl-2024-joint}; BEA (Building Educational Applications---19th iteration) \cite{bea-2024-innovative}; Argument Mining (11th iteration) \cite{argmin-2024}; and NLLP (Natural Legal Language Processing--3rd iteration) \cite{nllp-2024-1}, among others. Key objectives featured in these workshops include creating structured resources, increasing interpretability, evaluating models and resources, and coping with low-resource settings. This underscores the importance of interdisciplinary work between all of NLP, linguistics, and other fields from humanities and social sciences.

\section{Conclusion and Outlook}\label{sec:conclusion}

For a long time, automating analysis into linguistically-based representations---classic ``cL``---was undertaken in part with the aim of paving the path towards NLU, building modules that capture features of language based on human conceptualizations (realized as manually crafted rules, or latent rules induced through training on annotations).
Given the rise of LLMs that leverage huge amounts of data to recognize patterns in text, the question of the relationship between natural language processing and linguistics---in particular, how linguistic knowledge benefits work in natural language processing---is a pressing one.
In this paper, we have illustrated the role of linguistics when compiling resources and conducting system evaluations (RE-); when building systems in low-resource settings \mbox{(-L-)}; when pursuing granular interpretation or control of large-data systems \mbox{(-IE-)}; and when connecting NLP to the study of human language (-S). While explicit linguistic knowledge is not necessary to achieve high accuracy on all NLP tasks, in this work we have highlighted the engineering utility of language knowledge. 

Beyond continuing to develop cL resources and tools, our review has highlighted \textbf{other pertinent challenges for the NLP community}, including:
Efforts towards preserving the world's languages via cooperation between linguists, machine learning experts, and language communities (\textbf{R}\textbf{L}\textbf{S});
language learning applications and computational models of language acquisition, with potential for both the use and analysis of linguistic phenomena in NLP (\textbf{EL}\textbf{IES});
integration of symbolic representations and linguistic criteria generally into approaches that promote interpretability and explanation (\textbf{IE}).

\begin{acknowledgments}
Over the years, many scholars have influenced our perspectives on the relationship between NLP and Linguistics. We are grateful for feedback received in the course of writing this paper---from Julius Steen and Amir Zeldes on a draft, from Vivek Srikumar and Jena Hwang in a discussion, and Roland Kuhn on the problem of processing polysynthetic languages, as well as from reviewers. We are also thankful for feedback that was received during invited presentations of this work at the 2024 NFDI Text+ plenary in Mannheim, and in Sowmya Vajjala's NLP talk series. %Finally, we would like to thank anonymous reviewers for their thoughtful comments.
\end{acknowledgments}

\starttwocolumn
\bibliographystyle{compling}
\bibliography{relies}

\end{document}